\documentclass[10pt]{llncs}

\usepackage{times}
\usepackage{url}
\usepackage{latexsym}
\usepackage[latin9]{inputenc}
\usepackage{listings}
\usepackage{amsmath}
\usepackage{amssymb}
\usepackage{graphicx}
\usepackage{subscript}

\begin{document}

\title{What Kind of Programming Language Best Suits Integrative AGI?}

\author {Ben Goertzel }

\institute{SingularityNET Foundation and OpenCog Foundation }

\maketitle
\begin{abstract}
What kind of programming language would be most appropriate to serve the needs of integrative, multi-paradigm, multi-software-system approaches to AGI?   This question is broached via exploring the more particular question of how to create a more scalable and usable version of the  "Atomese" programming language that forms a key component of the OpenCog AGI design (an "Atomese 2.0") .  It is tentatively proposed that

\begin{itemize}
\item The core of Atomese 2.0 should be a very flexible framework of rewriting rules for rewriting a metagraph (where the rules themselves are represented within the same metagraph, and some of the intermediate data created and used during the rule-interpretation process may be represented in the same metagraph).    
\item This framework should (among other requirements)
\begin{itemize}
\item support concurrent rewriting of the metagraph according to rules that are labeled with various sorts of uncertainty-quantifications, and that are labeled with various sorts of types associated with various type systems.   A gradual typing approach should be used to enable mixture of rules and other metagraph nodes/links associated with various type systems, and untyped metagraph nodes/links not associated with any type system.   
\item allow reasonable efficiency and scalability, including in concurrent and distributed processing contexts, in the case where a large percentage of of processing time is occupied with evaluating static pattern-matching queries on specific subgraphs of a large metagraph (including a rich variety of queries such as matches against nodes representing variables, and matches against whole subgraphs, etc.)
\item allow efficient and convenient invocation and manipulation of external libraries for carrying out processing that is not efficiently done in Atomese directly
\end{itemize}
\item Among the formalisms we will very likely want to implement within this framework is {\it probabilistic dependent-linear-typed lambda calculus} or something similar, perhaps with a Pure IsoType approach to dependent type inheritance.  Thus we want the general framework to support reasonably efficient/convenient operations within this particular formalism, as an example.  
\end{itemize}

\end{abstract}

\section{Context and Motivations}

The history of AI has persistently featured fascinating feedback, synergy and tension between AI system design and programming language design.   Numerous researchers have come to the conclusion that, to make the radical AI advances they sought, they would require a better and more AI-friendly programming language environment.   Thus we got  languages like LISP and Prolog and their derivates.  Which have taught us a lot about AI and programming, yet without leading so far to the hoped-for AI breakthroughs.

Contemporary neural net based AI hasn't focused on introduction of new programming languages, but rather on new libraries such as Tensorflow, Torch, Theano and so forth.   On the other hand, the probabilistic programming paradigm has led to a remarkable profusion of new languages, most of which have arguably been unnecessary and distracted focus from the problem of efficiently executing probabilistic programs applied to real-world situations.

If one wants to pursue an integrative, multi-paradigm approach to AGI, then the situation as regards programming languages remains very far from optimal.   If one want to integrate, say, a logic programming system with a deep neural net perception system and a program learning system based on higher order functional types -- one is quite likely to want to implement the three components in different languages, and glue them together with scripts written in a simple language such as python.   Either that or one decides to value consistency and unity over elegance and efficiency, and shoehorns all three into a single language, reconciling oneself to either dramatic inefficiency or unwieldy, awkward code.

We have faced these issues recently in thinking through an envisioned redesign and reimplementation of the OpenCog AGI platform.  The current version of OpenCog relies heavily on a tool called the OpenCog Pattern Matcher, which is implemented in Scheme and is able to carry out highly complex procedure execution and predicate evaluation in the course of matching patterns against OpenCog's "Atomspace" weighted, labeled hypergraph knowledge store.   This Pattern Matcher is powerful but has become problematic for various reasons, including the lack of any built-in type system with an efficient type checker associated to it, and the complexity of interlacing the pattern matching process with calls to external processing tools such as deep neural net toolkits.    So we have begun designing a replacement we call "Atomese 2" -- Atomese being the informal name given to Scheme scripts that invoke Atomspace API calls and OpenCog Pattern Matcher queries.

It turns out that many of the conceptual and formal issues arising in the context of Atomese 2 design are of significantly broader importance, and are things that would arise in any attempt to create a programming language having both realistic efficiency and elegance in the context of integrative AGI applications.  In this paper we will review our thinking regarding Atomese 2, but keeping an eye always on the broader issues raised.   In the end what's important for AGI is not any specific programming language, but rather the underlying principles and structures, which may ultimately be implemented in a variety of different languages.

\section{Atomese 2 -- Conclusions and Considerations}

The OpenCog AGI framework, within which the Atomese language under discussion here operates, is centered on  a large, distributed, weighted labeled metagraph called the "Atomspace."  Atomese is then a custom language  specialized in pattern-matching and transforming this metagraph ("Atom" being OpenCog lingo for metagraph nodes or links).  

\begin{figure}
\begin{center}
\label{fig:jay-query}
\includegraphics[scale=0.8]{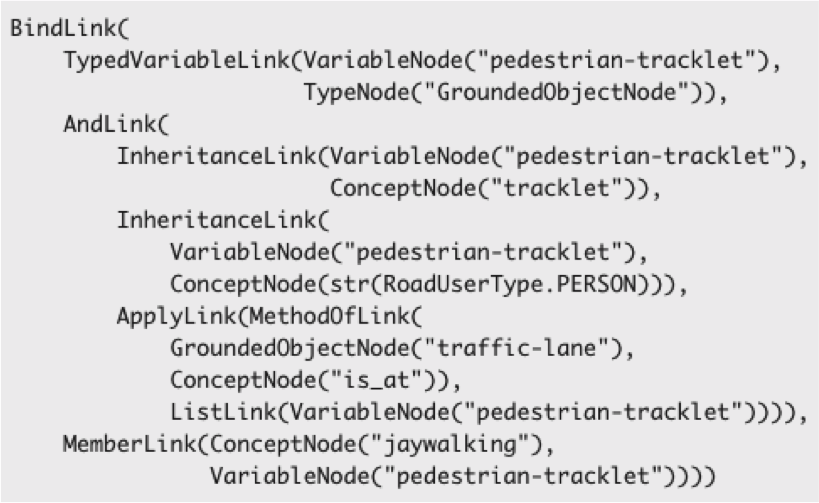}
\caption{Atomese expression that recognizes simple forms of jaywalking based on output of deep neural visual recognizers.  Application of the expression is mediated by OpenCog's URE rule engine, which leverages OpenCog Pattern Matcher internally.}
\end{center}
\end{figure}

Just to give a flavor, a simple example of Atomese 1 usage is given in Figure \ref{fig:jay-query} -- drawn from an application built by Cisco Systems \footnote{ \url{https://www.youtube.com/watch?v=s7EtRJatVmg}}  in collaboration with SingularityNET Foundation, applying OpenCog to fuse results from multiple vision-processing deep neural nets to make inferential judgments about street scenes.

\begin{figure}
\begin{center}
\label{fig:jay-image}
\includegraphics[scale=0.6]{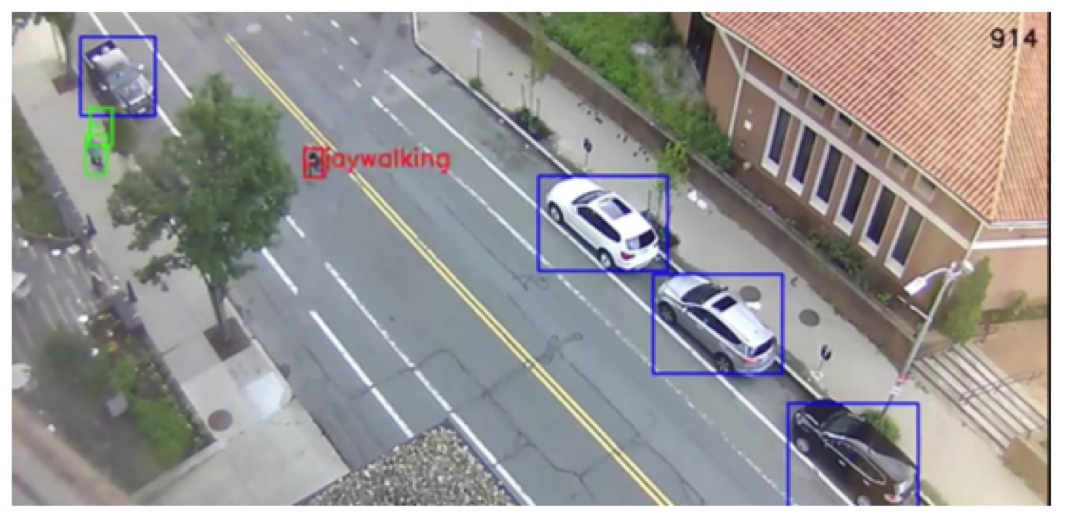}
\caption{Visual example of jaywalking that is recognized by Atomese expression in Figure \ref{fig:jay-query}}
\end{center}
\end{figure}

The OpenCog AGI design includes a carefully wrought combination of multiple AI methods such as probabilistic logical reasoning and pattern mining; probabilistic evolutionary program learning; neural net based attention allocation; neural-symbolic usage of deep neural nets for language, vision and sound; algorithmic chemistry based computational creativity ... and more.   It is an open-ended framework intended to allow experimentation with a variety of different AI algorithms and approaches.   On the other hand, the assemblage of AI tools already being explored and experimented with in an OpenCog context is sufficiently broad as to militate strongly toward an extremely flexible design.

Design of Atomese 2 becomes inextricably bound up with design of the overall OpenCog framework, including the Atomspace itself and the specific AI tools and methods to be implemented in Atomese and run in the context of the Atomspace.    Among the many issues that arise in this design process are:

\begin{enumerate}
\item What should the core Atomese formal language be?
\item Algorithmic approach to Atomese interpretation/compilation
\item Utilization of core Atomese to support various specialized formal languages useful for various AI algorithms
\item Surface form of Atomese language ("syntactic sugar")
\item RAM-based local Metagraph store -- which must be optimized for heavy Atomese usage of certain sorts
\item Distributed and persistent Metagraph store, perhaps with distributed RAM-based middleware as well
\item Atomese libraries corresponding to particular AI algorithms and approaches (e.g. the ones involved in OpenCog already)
\item Mode of integration of Atomese programs with external data/knowledge stores and processing and learning frameworks (e.g. external deep neural net libraries)
\end{enumerate}

\noindent In this paper we do not aim to address all these issues in depth, but rather focus on the first three.

In the context of core Atomese and its use to implement other formal languages (1 and 2 above), we need to think about issues such as:

\begin{itemize}
\item simple representation of the various Atom types in play in the OpenCog design
\item effective representation of external entities like knowledge-stores, specialized learning algorithms, simulations etc. as monads [or using some other powerful mode of encapsulation]
\item breaking down pattern-matching process into simple atomic operations like ``match this pattern at this location`` and ``move locus of pattern-matching to?"
\item compatibility of the above breakdown w/ concurrent and distributed processing
\item timed pattern matching should be fairly easily efficiently implementable 
\item How can we make it simple for a developer to add low level optimized support for some particular set of predicates / schema that's of interest?
\end{itemize}

\section{The Role of Static Pattern Matching in Atomese}

A peculiarity of the intended AGI use-case for Atomese is that we can assume the vast majority of processing time is spent on two key operations,

\begin{enumerate}
\item checking a particular (generally small) sub-metagraph to see if a certain pattern is matched there, {\it for a wide variety of patterns, to be dynamically generated and not foreseeable in advance}
\item applying a small set of metagraph rewrite rules to a particular (generally small) metagraph
\end{enumerate}

\noindent There can be assumed to be roughly comparable balance between these two sorts of operations.   

Also, there is a need both for rapid processing of these sorts of queries on a large metagraph in local RAM, and for distributed processing of these sorts of queries on a metagraph that is stored across numerous machines.

This means it is not important that Atomese be especially efficient at, say, sorting lists or computing the FFT.   What is important is that it is efficient at doing the above two operations and piping around, and doing simple manipulations on, the results of these operations.   If e.g. list-sorting or mathematical calculations are needed, it is assumed that Atomese will get these things done via referencing libraries coded in other languages.   Elegant and efficient interfacing with a variety of other languages and toolkits is thus highly important.

\subsection{Decomposing the Pattern Matching and Rule System Execution Process}

In OpenCog, the two key operations mentioned above are packaged up into the Pattern Matcher, which embodies a particular search algorithm and a variety of programming-language mechanisms along with basic pattern-matching functionality; and the Unified Rule Engine which executes a set of rules using forward or backward chaining, using the Pattern Matcher to manage rule application.   This is a powerful approach but also can be overly rigid.   

One design idea under discussion regarding Atomese 2 is that the interpretation process should break down an Atomese program into small chunks, which will mostly exemplify the two operations mentioned above (local pattern matching and local rewrite rules), plus operations of traversal within the metagraph.   Atomese scripts will then combine these chunks in various ways, dispatching some to remote machines as needed.   Improved versions of what the current OpenCog Pattern Matcher and URE do would then be implemented at Atomese scripts combining these elementary chunks.   In essence, in this approach Atomese scripts will  use functional programming constructs to interweave pattern matching with procedural content execution.

A few other particularities of pattern matching in an integrative AGI context are that:

\begin{itemize}
\item Static pattern matching must include matching against Atoms representing variables (i.e. variables must be first-class citizens, treated like any other cognitive content)
\item It must also include matching of individual query terms against sub-hypergraphs (not just individual nodes/links)
\end{itemize}

It should also be noted that static Atomspace pattern matching via Breadth-First-Search can be implemented so as to efficiently exploit multi-GPU architectures (using Gunrock \cite{wang2016gunrock} or similar tools).

\section{A Two-Layer Language Design}

This section outlines a potential high level approach to Atomese 2 design based on the above concepts.

\subsection{A Generic Atomese Core}

To enable the flexible exploration needed to work from our current state of knowledge toward a refined AGI design, the Atomese  core must be something quite generic -- e.g. it must comprise both

\begin{itemize}
\item a way of defining/manipulating Atoms (including specifying Atoms that embody rewriting rules for mapping sub-metagraphs into sub-metagraphs)
\item a way of defining/utilizing Atom type systems and Atom indexes associated w/ specific Atom types or type-systems.   (Note that the type systems defined should be defined within the same metagraph in which the Atoms reside.)
\end{itemize}

\noindent For each Atom type system that one defines, one should be able to plug in a type-checker / type-inference-system.

The Atomese  core may then need to be a rather generic gradual-typing framework, that deals with a system involving some Atoms that have incompletely specified or nonspecified types, and other Atoms that are defined w/in specific type systems.   

\subsubsection{Gradual Typing}

For background on gradual typing see: \cite{Siek2014} \cite{Siek2012}.  

While the matter seems not to have been explored theoretically in great detail, it seems intuitive that gradual typing in programming languages should map via Curry Howard type isomorphisms into paraconsistent logics of some sort.    

Achieving efficient execution of gradually typed languages is challenging (though not infeasible) because of obvious issues regarding casting between the dynamically and statically types parts of a program \cite{new_2018}.  However, given the peculiarities of Atomese, this bottleneck may not matter as the pattern-matching bottleneck may be more severe.

\subsection{Critical Formalizations Atop the Core}

The next larger layer of the onion would then be a specific type system (or small set of type systems) that we find to be interesting and potentially adequate for the particular AGI-oriented algorithms we're developing in practice.  This would be a set of languages / formal systems developed on top of core Atomese.   This is where, tentatively, it seems probabilistic linear dependent types will come in.

The obvious advantage of this sort of layered approach is that we can then modify the "probabilistic linear dependent types" or other specific formalizations a little later without having to rebuild the architecture.   However, we should expect that in practice nearly all users are going to end up working with the "specific type system" layer of the onion we initially create, rather than the "generic gradual-typing based Atom  and Atom-type-system framework" layer.

\section{Some Specific Type Systems of Apparent AGI Relevance}

One hypothesis that seems very much worth exploring is to use {\it probabilistic linear dependent types with IsoType type inference} as a formalization on top of Atomese core, with power to drive both probabilistic logic and also related applications such as probabilistic program learning.

It seems that this particular flavor of type system may meet the needs of a variety of AI algorithms currently existing in OpenCog, plus others that have been proposed for OpenCog integration: Probabilistic Logic Networks, surprisingness-based pattern  mining, probabilistic evolutionary program learning (MOSES), probabilistic programming (including cases with neural nets or probabilistic logic inference on the back end), nonlinear-dynamical attention allocation, content-addressable episodic memory, neural-symbolic perception processing and action control.

The full argument why this particular formalization direction is valuable for meeting these needs of these AI algorithms is involved with many parts and would be too lengthy to full elaborate here.  Rather, here only a few of the more critical points will be sketched.

\subsection{Dependent Types}

Dependent types are valuable in an AGI context because they enable elegant manifestation of the morphism between declarative knowledge (logic expressions) and procedural knowledge (programs).   Programs expressed with dependent types can be very straightforwardly interpreted as logic expressions.   Converting between procedural and declarative knowledge is key to AGI, and having a formalism that makes this convenient is high value.   An elegant prototype interpreter for lambda calculus with dependent types is Lambda-Pi, available as open source code on Github \footnote{ \url{https://github.com/lambda-pi-plus/lambda-pi-plus} , \url{https://github.com/tdietert/lambda-pi}}.

\subsection{IsoType Systems}

Pure IsoType Systems (PITS) are a way to  get (a lot of) the power of dependent types without making type-checking undecidable \cite{yang_oliveira_2019}.   They may also help with making dependent type checking not only decidable but reasonably fast, though this is still an active research topic.  Whether their limitations are important from an AGI perspective is not clear.

\subsection{Linear Types}

{\it Engineering General  Intelligence} \cite{EGI1} \cite{EGI2}, the foundational book outlining the theory behind OpenCog, has a whole section on "effort management" -- counting the computational resource usage of each cognitive operation and using this in planning etc.   This is important and ties into Occam's Razor heuristics which are key to AGI theory. though we haven't dealt with explicit effort management much in our practical OpenCog work so far.

Linear logic basically lets you count resource usage in the guts of your logic engine (or equivalently, your program execution process).  Of course there are always other ways to do this, but having it built into the logic is a way  that fits naturally with reflection and meta-computation.  Dependent types have been gotten to work with linear types \cite{benton_2015}; and pattern matching with linear types has also been explored \cite{Anders_2010}.

We note that to make probabilistic linear lambda calculus confluent you choose either call-by-value or call-by-reference.   Similarly making either of these choices renders type checking decidable in dependent type theory w/ isotypes.   One guesses that making probabilistic linear lambda calculus with dependent linear types, if one wants to restrict type equivalency to isotyping, then one will get both confluence and decidability from either choice of call-by-reference or call-by-value.

\subsection{Probability/Logic Interoperation}

Probabilistic methods are probably the biggest innovation in AI over the last couple decades, and it seems clear that including probabilistic representation and manipulation at the basic level is going to be a good idea for any AGI engine.

Recent work  \cite{Faggian2019} gives a variant of probabilistic lambda calculus that is confluent (it achieves confluence by limiting the reductions that can take place, in a manner framed via linear logic).

This sort of low-level probability/logic integration lays the groundwork for specific probabilistic-logic math aimed at deductive, inductive, abductive and other forms of inference, such as e.g. OpenCog's Probabilistic Logic Networks (PLN) framework \cite{PLN} carries out.

PLN depends heavily on non-confluent reductions in probabilistic logic expressions, however these are necessarily going to be kind of heuristic and history-guided, so it makes sense for them to live in the next layer of the onion -- i.e. we have core Atomese, then probabilistic linear dependently types lambda calculus or similar built on that, then PLN built on that.   But the building of PLN on top of elaborated lambda calculus can use the same basic Atomese syntax and interpreter as the building of elaborated lambda calculus on core Atomese.

\section{Toward an Integrative AGI Language and Architecture}

\begin{figure}
\begin{center}
\label{fig:TrueAGI-arch}
\includegraphics[scale=0.45]{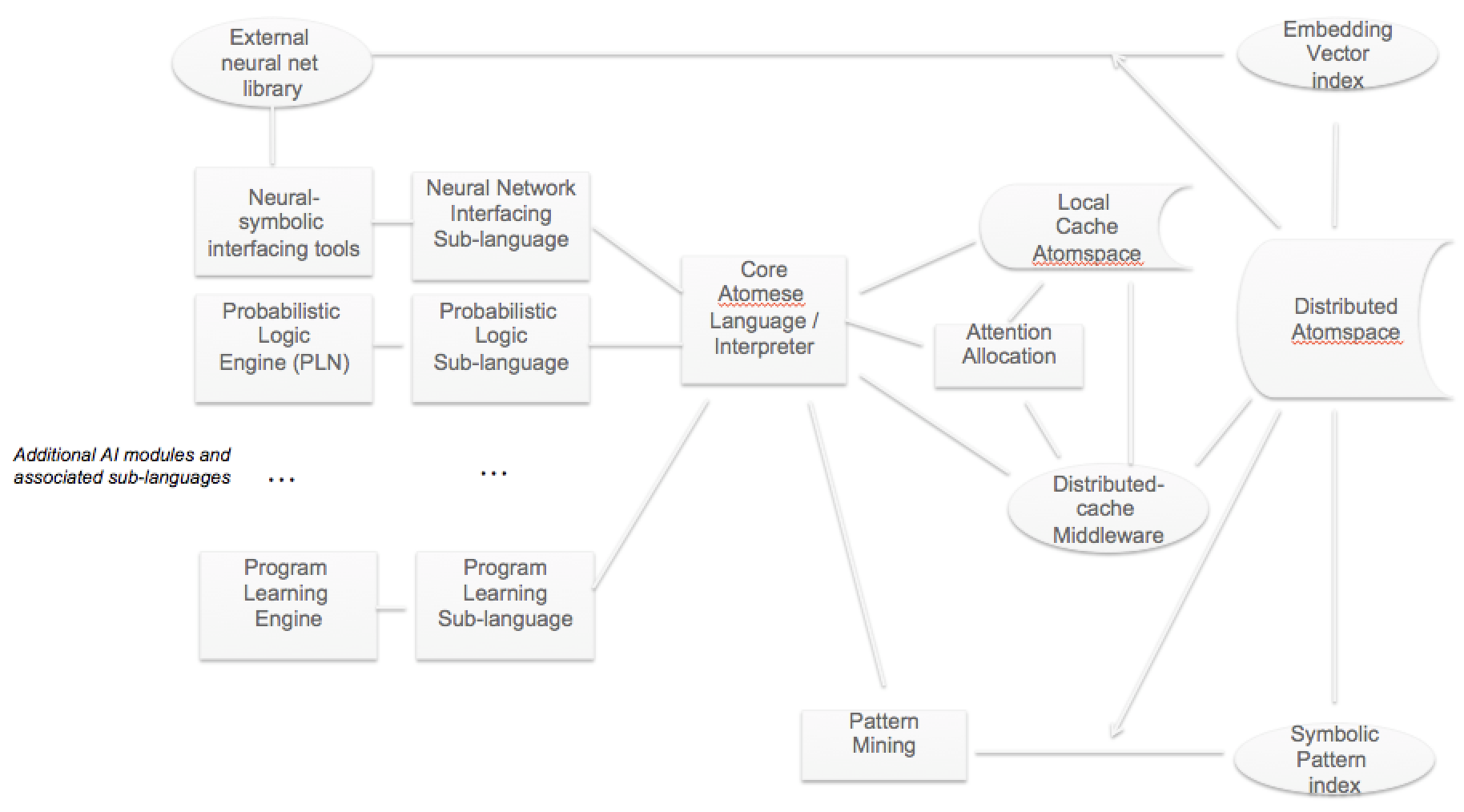}
\caption{A software architecture for integrative AGI, with a pattern-matching-focused, gradually typed Atomese language at the core.}
\end{center}
\end{figure}

Figure \ref{fig:TrueAGI-arch} summarizes the overall "next-generation OpenCog" architecture that is suggested by the above thoughts on Atomese 2 design.

In the context of the above figure, e.g. PLN logic might end up using a type system founded in probabilistic linear dependent types with IsoType type inference.   On the other hand, for automated program learning it might be decided that the IsoType approach is too restrictive, and it's better to bite the inefficiency bullet a little harder and go with a more flexible type inheritance mechanism.   

In this case, via the gradual typing approach, we could have some Atoms that are not typed at all, and can thus play a role in either the PLN or program learning focused type systems.  On the other hand, if program learning generates a program that then needs to be reasoned about, this will necessitate a mapping from the program-learning type system to the probabilistic-logic type system.   There will be some equations that are consistent in one of these logics but not the other (in particular, perhaps some that are consistent using IsoTypes and not using more flexible inheritance mechanisms) -- thus rendering the overall framework paraconsistent, rather than strictly consistent.

\section{Conclusion}

There is much more to be learned here and we are in the middle rather than at the end of the Atomese 2 language design process.   However, the thinking we've done so far has already highlighted some issues of likely broader relevance in the context of integrative approaches to AGI.  For instance, the dominance of RAM-based pattern matching in terms of runtime resource consumption, and the convenience of a gradual typing approach, are points going well beyond the particulars of OpenCog's chosen assemblage of AI algorithms.

Formulating the right programming language is very unlikely to magically produce a workable AGI system.   However, a programming language and environment that eases rapid implementation and scalable deployment of cross-paradigm AI algorithmics, could certainly dramatically accelerate progress.

\section*{Acknowledgements}  Many of the ideas reviewed here originated in discussions with Alexey Potapov, Cassio Pennachin, Vitaly Bogdanov and other SingularityNET colleagues -- though the specific presentation of these ideas here is my own responsibility for better and/or worse.

\bibliographystyle{splncs03} 
\bibliography{bbm}








\end{document}